\documentclass[letterpaper]{article} 
\usepackage{aaai2026}  
\usepackage{times}  
\usepackage{helvet}  
\usepackage{courier}  
\usepackage[hyphens]{url}  
\usepackage{graphicx} 
\usepackage{natbib}  
\usepackage{caption} 
\usepackage{algorithm}
\usepackage{algorithmic}
\usepackage{amsmath}
\usepackage{booktabs}
\usepackage{tabularx}
\usepackage{xcolor}
\usepackage{lipsum}
\usepackage{pifont}
\usepackage{multirow}

\newcommand{\cmark}{\ding{51}}%
\newcommand{\xmark}{\ding{55}}%

\usepackage{xcolor}
\newcommand{\answerYes}[1]{\textcolor{blue}{#1}} 
\newcommand{\answerNo}[1]{\textcolor{teal}{#1}} 
\newcommand{\answerNA}[1]{\textcolor{gray}{#1}}

\usepackage{newfloat}
\usepackage{listings}
\DeclareCaptionStyle{ruled}{labelfont=normalfont,labelsep=colon,strut=off} 
\floatstyle{ruled}
\newfloat{listing}{tb}{lst}{}
\floatname{listing}{Listing}
\title{RoIt-XMASA: Multi-Domain Multilingual Sentiment Analysis Dataset for Romanian and Italian}
\author{
    Andrei-Marius Avram,
    Aureliu-Valentin Antonie\thanks{Equal contribution. Alphabetical author order.},
    Cosmin-Mircea Croitoru\footnotemark[1],
    Vlad-Andrei Muntean\footnotemark[1],
    Dumitru-Clementin Cercel\thanks{Corresponding author.}
}
\affiliations{
    National University of Science and Technology POLITEHNICA Bucharest,\\
    Bucharest, Romania\\
    dumitru.cercel@upb.ro
}

\usepackage{bibentry}

\begin{document}

\maketitle

\begin{abstract}
We present RoIt-XMASA, a multilingual dataset that extends the Cross-lingual Multi-domain Amazon Sentiment Analysis to Italian and Romanian, comprising 36,000 labeled reviews across three domains (books, movies, and music) and 202,141 unlabeled samples. To address cross-lingual and cross-domain challenges, we propose a multi-target adversarial training framework that employs loss reversal with meta-learned coefficients to dynamically balance sentiment discrimination with domain and language invariance. XLM-R achieves an F1-score of 66.23\% with our approach, outperforming the baseline by 4.64\%. Few-shot evaluation shows that Llama-3.1-8B achieves 58.43\% F1-score, revealing a meaningful trade-off between the efficiency of prompting-based approaches and the higher performance of task-specific fine-tuning.
\end{abstract}


\section{Introduction}

Cross-domain and cross-lingual sentiment analysis remains a fundamental challenge, requiring models to generalize across both linguistic and topical boundaries. Although multilingual models have advanced significantly, the double shift in language and domain, compounded by the scarcity of resources for languages such as Romanian, presents a major obstacle \cite{zhao2024systematic}. These limitations emphasize the need for frameworks that leverage large-scale unlabeled data and adversarial learning \cite{ganin2016domain,chen2018adversarial} to extract invariant sentiment features across disparate distributions.

We introduce RoIt-XMASA, a large-scale multilingual dataset extending the Cross-lingual Multi-domain Amazon Sentiment Analysis (XMASA) corpus \cite{blitzer2007biographies} to Italian and Romanian. Our dataset comprises 36,000 annotated reviews equally distributed across three domains (books, movies, and music) and two languages, with an additional 202,141 unlabeled samples for semi-supervised learning. Each review is annotated with ratings from 1 to 5 stars (excluding 3), maintaining the original XMASA structure while adapting to the linguistic characteristics of Italian and Romanian.

Our methodological contribution introduces a multi-target adversarial training framework \cite{ganin2016domain,chen2018adversarial} that simultaneously optimizes for sentiment classification while learning representations invariant to domain and language attributes. We employ loss reversal \cite{avram2024histnero} rather than gradient reversal and dynamically adjust adversarial coefficients through meta-learning \cite{vettoruzzo2024advances}, thus eliminating the need for manual hyperparameter tuning.

Experiments with multilingual variants of BERT \cite{devlin2019bert} demonstrate that our approach obtains substantial improvements: XLM-R \cite{conneau2020unsupervised} achieves an F1-score of 66.23\% (an improvement of 4.64\% over the baseline) when using both domain and language as adversarial targets. We also establish baselines using recent open-source large language models (LLMs), with Llama-3.1-8B \cite{dubey2024llama} reaching a 58.43\% F1-score in few-shot settings. 

Furthermore, we investigate the utility of the 202,141 unlabeled samples in RoIt-XMASA for unsupervised domain adaptation. By performing a computationally efficient fine-tuning phase with Low-Rank Adaptation (LoRA) \cite{hulora} for a single epoch, we assess how brief exposure to domain-specific unlabeled data in Romanian and Italian can bridge the gap between general-purpose pre-training and the specific linguistic registers of e-commerce reviews. This adaptive step aims to improve the model alignment with the target distributions before the final classification task.

The main contributions of our work can be summarized as follows:
\begin{itemize}
\item We present RoIt-XMASA, the first large-scale extension of the XMASA dataset \cite{blitzer2007biographies} to Italian and Romanian, providing 36,000 labeled and 202,141 unlabeled reviews balanced across domains and languages\footnote{The dataset is available at the following link: \url{https://huggingface.co/datasets/avramandrei/RoIt-XMASA}}.
\item We propose a multi-target adversarial training framework with meta-learned coefficients that achieves consistent improvements across all tested multilingual models.
\item We establish comprehensive baselines using both fine-tuned encoder models and few-shot LLM approaches, revealing distinct linguistic patterns between Italian and Romanian reviewing styles with implications for cross-cultural natural language processing  (NLP) applications.
\item We validate the utility of the unlabeled subset by demonstrating that an unsupervised adaptation through LoRA measurably improves LLM performance, effectively aligning models with target languages and domains.
\end{itemize}

\section{Related Work}
\label{app:related_work}

\subsection{Evolution of Multilingual Sentiment Analysis}
Multilingual sentiment analysis has evolved significantly from early methods that relied on machine translation \cite{araujo2016evaluation} and lexicon-based features \cite{qi2023sentiment}. Initial cross-lingual approaches often involved translating test data into a high-resource language such as English (a technique known as a translate-test) or training classifiers on translated source data \cite{prettenhofer2010cross,wan2009co}. Another line of work focused on creating multilingual sentiment-aware word embeddings, either by aligning monolingual vector spaces or learning them jointly from parallel corpora, sometimes using pivot languages to bridge resource gaps \cite{xu2017towards}. We note that these methods frequently suffer from translation errors and the loss of cultural nuances.

The advent of pre-trained language models marked a paradigm shift in the field. Multilingual models such as mBERT \cite{devlin2019bert} and XLM-R \cite{conneau2020unsupervised} have demonstrated a remarkable capacity for zero-shot cross-lingual transfer by learning shared representations in more than 100 languages. Subsequent research has extensively benchmarked these models, confirming their effectiveness and also highlighting performance disparities across languages and tasks \cite{rajda2022assessment, augustyniak2023massively}. Models are typically evaluated as static feature extractors for a downstream classifier or via full fine-tuning, with the latter generally obtaining superior performance but at a higher computational cost.

Despite the success of large-scale multilingual models, significant challenges remain, particularly for low-resource languages and culture-dependent tasks. For example, sentiment expression is deeply intertwined with cultural context, and models pre-trained on generic web corpora may fail to capture fine-grained subtleties \cite{augustyniak2023massively}. This has spurred efforts to create high-quality and specialized datasets for underrepresented languages, such as the NusaX corpus for Indonesian dialects \cite{winata2023nusax}. Such resources are critical for developing models that are not only linguistically competent but also culturally aware, motivating the creation of our dataset, RoIt-XMASA.

\subsection{Cross-Lingual Adaptation and Robustness}
To improve the model robustness across languages and domains, researchers have explored adversarial training techniques \cite{chen2018adversarial}. The core idea is to learn feature representations that are discriminative for the primary task (e.g., sentiment classification) but invariant to nuisance variables like the language or domain of the input text. This is often achieved by introducing a domain classifier that is trained to predict the nuisance variable from the learned features, while the main feature extractor is trained to fool this classifier, typically through a gradient reversal layer \cite{ganin2016domain,ye2020feature}. This forces the model to learn more generalized and transferable features, making it a powerful technique for cross-lingual and cross-domain adaptation.

Most recently, the focus has shifted toward LLMs and their ability to perform sentiment tasks via prompting or instruction tuning. Although models like Llama and Qwen exhibit strong zero-shot capabilities, their performance in specialized domains or low-resource languages often benefits from targeted adaptation \cite{hulora}. Parameter-efficient techniques, particularly LoRA, have emerged as a standard for adapting these massive models without the prohibitive costs of full-parameter updates. Recent studies suggest that performing an initial phase of unsupervised domain adaptation—where the model is exposed to unlabeled target-domain text—can significantly improve alignment and subsequent classification accuracy \cite{saad2023udapdr,zhang2024gongbu,xing2025multi}. Our work builds on this by quantifying the impact of such unlabeled adaptation on the Romanian and Italian review domains, positioning it as a complementary strategy to adversarial training in multi-domain scenarios.

\section{RoIt-XMASA Dataset}

RoIt-XMASA, the multilingual dataset we introduce in this work, extends the XMASA dataset \cite{blitzer2007biographies} to Italian and Romanian languages, maintaining the three-domain structure (books, movies, and music) while adapting to the linguistic and cultural characteristics of these languages. Our dataset contains 36,000 annotated reviews crawled from the Internet, which were equally distributed across all dimensions.

\subsection{Dataset Collection}

The construction of the corpus followed a language-specific acquisition strategy to ensure high-quality and authentic data. The Italian reviews were collected from Amazon to maintain consistency with the original source of XMASA. In contrast, due to the lack of Romanian-language reviews on Amazon, the Romanian reviews were acquired from a combination of representative local platforms and manually translated English sources to better capture native linguistic and cultural patterns for each of the three domains: (1) book reviews were collected from Goodreads\footnote{https://www.goodreads.com/} and Audiotribe\footnote{https://audiotribe.ro}, with additional samples manually translated from XMASA's English data to reach the target domain size of 6k reviews, (2) movie reviews were gathered from Cinemagia\footnote{https://www.cinemagia.ro/}, a prominent Romanian film review platform, and (3) music reviews were obtained from XMASA's English data and manually translated into Romanian.

The dataset follows a balanced design with 6,000 labeled samples in each of the train, validation, and test splits. Table \ref{tab:stats} presents the distribution across domains and languages. Each language-domain combination contains exactly 2,000 samples per split, ensuring balanced representation for both cross-lingual and cross-domain evaluation scenarios.

In addition to the labeled splits, we provide an unlabeled dataset of 202,141 samples to facilitate research on semi-supervised learning and domain adaptation. The unlabeled data set spans all three domains: Italian samples include 14,722 books, 32,970 movies, and 53,662 music reviews, while Romanian samples comprise 13,429 books, 60,023 movies, and 27,335 music reviews.

\begin{table}
\centering

\begin{tabular}{|l|cc|cc|cc|}
\toprule
\multirow{2}{*}{\textbf{Labeled Sets}} & \multicolumn{2}{c|}{\textbf{Books}} & \multicolumn{2}{c|}{\textbf{Movies}} & \multicolumn{2}{c|}{\textbf{Music}} \\
\cmidrule{2-7}
& \textbf{IT} & \textbf{RO} & \textbf{IT} & \textbf{RO} & \textbf{IT} & \textbf{RO} \\
\midrule
Train & 2k & 2k & 2k & 2k & 2k & 2k \\
Valid & 2k & 2k & 2k & 2k & 2k & 2k \\
Test & 2k & 2k & 2k & 2k & 2k & 2k \\
\midrule
\textbf{Total} & \textbf{6k} & \textbf{6k} & \textbf{6k} & \textbf{6k} & \textbf{6k} & \textbf{6k} \\
\bottomrule
\end{tabular}

\caption{It-XMASA and Ro-XMASA dataset statistics by domain and language on the train, validation, and test subsets.}
\label{tab:stats}
\end{table}

\subsection{Rating Distribution}

\begin{figure*}[!ht]
\centering
\includegraphics[width=\textwidth]{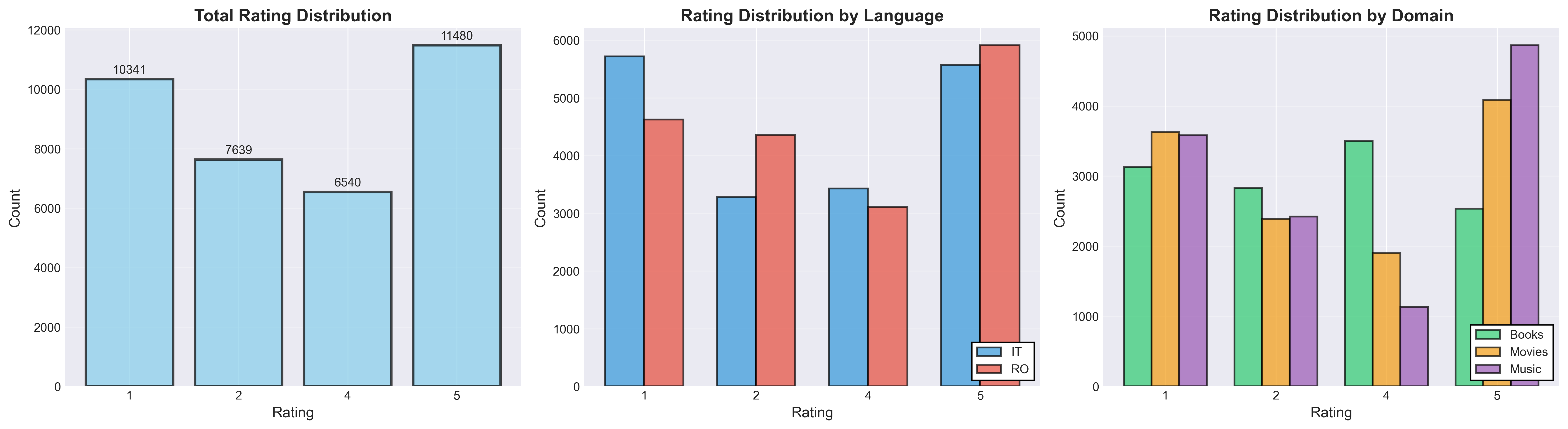}
\caption{Rating distribution across the RoIt-XMASA dataset. Left: overall rating distribution; Center: rating distribution by language; Right: rating distribution by domain.}
\label{fig:rating_dist}
\end{figure*}

Figure~\ref{fig:rating_dist} illustrates the rating distribution across the entire RoIt-XMASA dataset and its breakdown by language and domain. The dataset exhibits a naturalistic distribution with peaks at the extreme ratings (1 and 5 stars), reflecting the tendency of users to review products they either strongly like or dislike.

The distribution shows 10,341 one-star ratings, 7,639 two-star ratings, 6,540 four-star ratings, and 11,480 five-star ratings. This polarization pattern is consistent across both languages, with Italian showing a slightly more balanced distribution of intermediate ratings than Romanian. This pattern also appears in both the movies and the music domains.

\subsection{Textual Characteristics}

\begin{figure*}
\centering
\includegraphics[width=\textwidth]{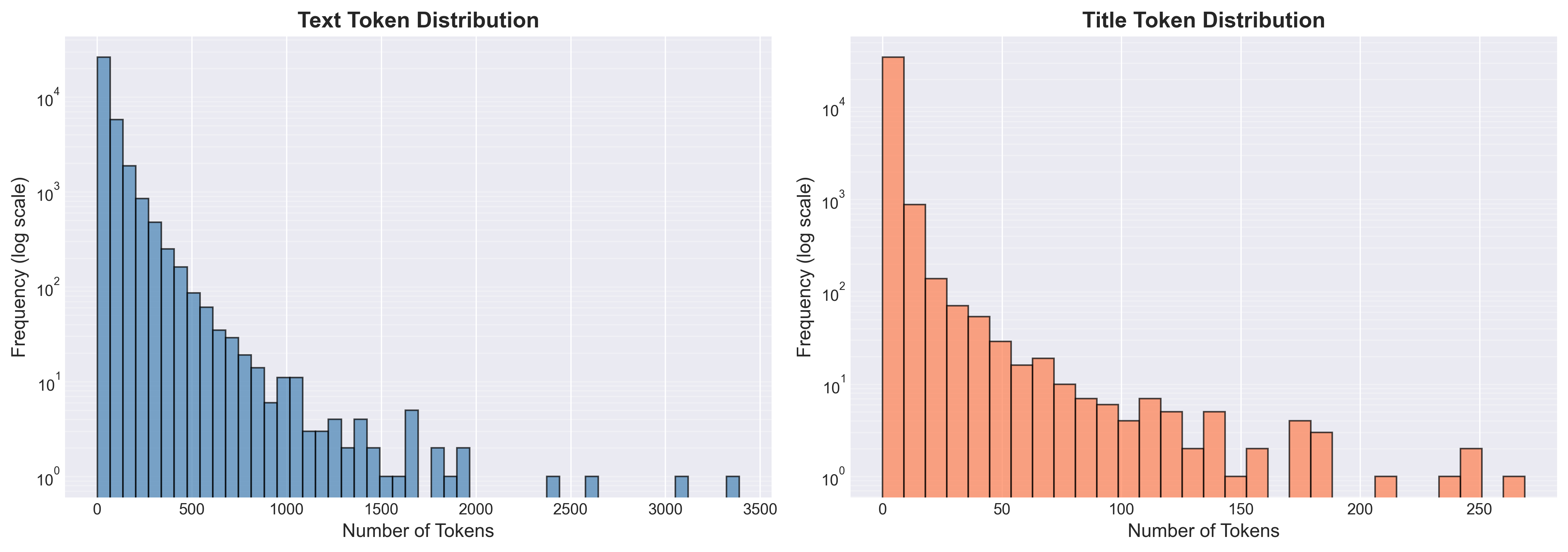}
\caption{Token distribution histograms for the text and title fields.}
\label{fig:token_dist}
\end{figure*}

The RoIt-XMASA dataset exhibits some variation in the review length and linguistic properties. Overall, reviews contain a mean of 63.18 tokens and a median of 33, with a rather high variance (i.e., a standard deviation of 100.78), ranging from single-token reviews to extensive critiques of up to 3,391 tokens. This wide distribution, illustrated in Figure~\ref{fig:token_dist}, reveals the diversity of user expressions in sentiment communication, with a characteristic long-tailed distribution typical of user-generated content.

The titles show even greater variability, with a mean of 2.20 tokens and a median of approximately 1 token, but a highly skewed distribution. The presence or absence of titles appears to be strongly influenced by cultural and platform-specific conventions, as revealed by differences across languages (i.e., out of the 36k collected samples, only 19,762 have titles). A comprehensive cross-language and cross-domain analysis of these textual characteristics is provided in the Appendix.

\subsection{Data Quality}

Quality assessment reveals that there are no duplicate entries in RoIt-XMASA when considering the combination of title and text fields. This ensures that each sample represents a unique review instance, preventing data leakage between splits and maintaining the integrity of experimental evaluations.
The RoIt-XMASA dataset underwent rigorous validation and cleaning procedures to ensure data quality:

\begin{itemize}
\item \textbf{Language verification}: All reviews were validated using the \texttt{cld3}\footnote{https://github.com/google/cld3} library to ensure correct language assignment. Reviews with confidence scores below 0.95 or that were detected as different languages were manually reviewed, corrected, or removed.
\item \textbf{HTML and special character removal}: Systematic cleaning removed all HTML tags, entities (e.g., \&amp;, \&nbsp;), and hidden control characters that could interfere with model processing.
\item \textbf{Text normalization}: Applied comprehensive normalization rules to standardize text representation while preserving semantic content. The complete normalization pipeline is detailed in the Appendix.
\end{itemize}

To validate the automated processing pipeline, we conducted manual verification by human annotators. Three native speakers for each language reviewed 100 randomly sampled reviews (200 total), assessing language correctness, sentiment-rating alignment, and text quality. The inter-annotator agreement achieved a Krippendorff's alpha of 0.82, indicating strong agreement and confirming the reliability of our data quality measures.

The balanced structure of the RoIt-XMASA dataset, the natural rating distribution, and various textual characteristics make it well-suited for investigating the challenges of sentiment analysis in both cross-lingual and cross-domain settings, extending the foundational work of \citet{blitzer2007biographies} to new linguistic territories while maintaining methodological consistency.

\section{Methodology}

\subsection{Multi-Adversarial Objective with Meta-Learned Coefficients}

We build on the multi-target adversarial training framework introduced by \citet{avram2025morovoc} and adapt it for sentiment classification. Let $f_{\theta}$ denote the shared encoder, and $h_{\phi}$ the task-specific classification head with task loss $\mathcal{L}_{\text{task}}$ for each of the three possible prediction objectives (i.e., rating, language, and domain). Then, in our setup, the rating prediction objective is designated as the primary task to be optimized, while the remaining two are treated as adversarial objectives.

Unlike previous work \cite{ganin2016domain,chen2018adversarial}  that employs gradient reversal, we adopt loss reversal, which directly flips the sign of adversarial objectives during optimization, shown in previous work to yield better results \cite{avram2024histnero,avram2025rolargesum}. Thus, the overall loss is as follows:

\begin{equation}
\mathcal{L} = \mathcal{L}_{\text{rating}} - \lambda_1 \mathcal{L}_{\text{domain}} - \lambda_2 \mathcal{L}_{\text{lang}},
\end{equation}
where $\lambda_1$ and $\lambda_2$ are adversarial coefficients. To avoid manual tuning, we adopt a meta-learning strategy: coefficients are updated to minimize a validation-set meta-loss:

\begin{equation}
\lambda \leftarrow \lambda - \eta \nabla_{\lambda} \mathcal{L}_{\text{meta}}(\lambda),
\end{equation}
with $\eta$ denoting the meta-learning rate. This procedure dynamically balances the invariance to adversarial attributes with the discriminability of the primary task.
 
All multilingual encoder models, namely M-BERT \cite{devlin2019bert}, XLM \cite{conneau2019cross}, and XLM-R \cite{conneau2020unsupervised}, were fine-tuned using the same training protocol to ensure a fair comparison across architectures. This includes identical optimization settings, batch sizes, learning rate schedules, and early stopping criteria. We report the complete set of hyperparameters used to fine-tune multilingual encoder models in the Appendix.

\begin{table*}[!ht]
    \centering
    \small
    \resizebox{\textwidth}{!}{
    \begin{tabular}{|l|c|c|cc|cc|cc|cc|cc|cc|}
    \toprule
    \multirow{2}{*}{\textbf{Model}} & \multirow{2}{*}{\textbf{Dom.}} & \multirow{2}{*}{\textbf{Lang.}} & \multicolumn{2}{c|}{\textbf{Books}} & \multicolumn{2}{c|}{\textbf{Movies}} & \multicolumn{2}{c|}{\textbf{Music}} & \multicolumn{2}{c|}{\textbf{IT}} & \multicolumn{2}{c|}{\textbf{RO}} & \multicolumn{2}{c|}{\textbf{Avg.}} \\
    \cmidrule{4-15}
    & & & \textbf{Acc.} & \textbf{F1} & \textbf{Acc.} & \textbf{F1} & \textbf{Acc.} & \textbf{F1} & \textbf{Acc.} & \textbf{F1} & \textbf{Acc.} & \textbf{F1} & \textbf{Acc.} & \textbf{F1} \\
    \midrule
    M-BERT & \xmark & \xmark & 62.15 & 57.23 & 65.82 & 60.45 & 63.05 & 57.64 & 66.92 & 62.18 & 60.42 & 54.70 & 63.67 & 58.44 \\
    M-BERT & \cmark & \cmark & 65.23 & 60.89 & 68.45 & 64.28 & \textbf{67.91} & 63.15 & 69.45 & 65.12 & \textbf{65.28} & 61.35 & 66.45 & 62.17 \\
    \midrule
    XLM & \xmark & \xmark & 60.34 & 57.12 & 63.78 & 60.89 & 60.89 & 57.79 & 64.23 & 61.45 & 59.11 & 55.75 & 61.67 & 58.60 \\
    XLM & \cmark & \cmark & \textbf{66.78} & 62.41 & 65.12 & 62.18 & 61.78 & 59.55 & 66.34 & 62.67 & 62.78 & \textbf{61.42} & 64.56 & 61.38 \\
    \midrule
    XLM-R & \xmark & \xmark & 63.18 & 60.34 & 66.82 & 63.78 & 63.71 & 60.65 & 67.45 & 64.23 & 61.69 & 58.95 & 64.57 & 61.59 \\
    XLM-R & \cmark & \cmark & 66.45 & \textbf{64.12} & \textbf{71.12} & \textbf{68.61} & 67.15 & \textbf{65.78} & \textbf{72.42} & \textbf{69.40} & 63.39 & 60.93 & \textbf{68.91} & \textbf{66.23} \\
    \bottomrule
    \end{tabular}
    }
    \caption{Multi-adversarial training results on RoIt-XMASA. Domain and language columns show performance by domain (Books, Movies, Music) and language (IT=Italian, RO=Romanian).}
    \label{tab:finetune_results}
\end{table*}

\subsection{Open-Source LLM Baselines}

We evaluate recent open-source large language models as baselines for cross-domain and cross-lingual sentiment classification. The selection criterion was to include models between 6B and 9B parameters, always using the latest version released for each family to ensure a representative and unbiased comparison. Concretely, we benchmark the few-shot performance of four open-source LLMs: Llama-3.1-8B\footnote{\url{https://huggingface.co/meta-llama/Llama-3.1-8B}}, Qwen3-8B\footnote{\url{https://huggingface.co/Qwen/Qwen3-8B}}, Mistral-7B-v0.3\footnote{\url{https://huggingface.co/mistralai/Mistral-7B-Instruct-v0.3}}, and DeepSeek-R1-8B\footnote{\url{https://huggingface.co/deepseek-ai/DeepSeek-R1-Distill-Llama-8B}}. 

Furthermore, to better align open-source models with the specific linguistic and domain distributions of our corpus, we performed an initial stage of fine-tuning on the unlabeled data. Using the 202,141 unlabeled samples available in RoIt-XMASA, we fine-tuned each LLM for one epoch using LoRA. This adaptation phase employed a causal language modeling objective, allowing the models to learn the nuances of Italian and Romanian review styles across the three domains without the computational overhead of full-parameter tuning.

All models were evaluated in a prompting-based setup, with prompts and LLM experiment hyperparameters detailed in the Appendix.

\section{Results}
\label{sec:results}

\begin{table*}[!ht]
    \centering
    \small
    \resizebox{\textwidth}{!}{
    \begin{tabular}{|l|c|cc|cc|cc|cc|cc|cc|}
    \toprule
    \multirow{2}{*}{\textbf{Model}} & \multirow{2}{*}{\textbf{\# Shots}} & \multicolumn{2}{c|}{\textbf{Books}} & \multicolumn{2}{c|}{\textbf{Movies}} & \multicolumn{2}{c|}{\textbf{Music}} & \multicolumn{2}{c|}{\textbf{IT}} & \multicolumn{2}{c|}{\textbf{RO}} & \multicolumn{2}{c|}{\textbf{Avg.}} \\
    \cmidrule{3-14}
    & & \textbf{Acc.} & \textbf{F1} & \textbf{Acc.} & \textbf{F1} & \textbf{Acc.} & \textbf{F1} & \textbf{Acc.} & \textbf{F1} & \textbf{Acc.} & \textbf{F1} & \textbf{Acc.} & \textbf{F1} \\
    \midrule
    Llama-3.1-8B & 0 & 58.23 & 49.12 & 62.45 & 53.68 & 60.08 & 50.08 & 63.12 & 54.23 & 57.34 & 47.69 & 60.25 & 50.96 \\
    Llama-3.1-8B & 5 & 61.89 & 56.78 & \textbf{66.23} & \textbf{61.85} & 63.12 & 58.15 & \textbf{66.34} & 60.89 & 61.15 & 56.63 & \textbf{63.66} & \textbf{58.43} \\
    \midrule
    Qwen3-8B & 0 & 56.45 & 48.23 & 60.34 & 52.78 & 58.05 & 49.26 & 61.12 & 52.89 & 55.44 & 47.29 & 58.28 & 50.09 \\
    Qwen3-8B & 3 & 59.67 & 54.23 & 61.45 & 56.12 & \textbf{63.78} & \textbf{59.34} & 63.45 & 57.92 & 59.82 & 55.11 & 60.11 & 54.97 \\
    \midrule
    Mistral-7B-v0.3 & 0 & 52.34 & 46.12 & 56.23 & 50.12 & 53.97 & 47.85 & 56.78 & 50.23 & 51.58 & 45.83 & 54.18 & 48.03 \\
    Mistral-7B-v0.3 & 5 & 53.89 & 47.23 & 57.45 & 51.34 & 54.56 & 48.40 & 57.89 & 51.12 & 53.71 & 46.86 & 55.30 & 48.99 \\
    \midrule
    DeepSeek-R1-8B & 0 & 59.78 & 54.89 & 63.67 & 58.67 & 61.29 & 55.49 & 64.45 & 58.92 & 58.71 & 53.78 & 61.58 & 56.35 \\
    DeepSeek-R1-8B & 1 & \textbf{62.45} & \textbf{57.81} & 63.34 & 59.12 & 61.23 & 55.67 & 65.12 & \textbf{61.23} & \textbf{61.89} & \textbf{57.04} & 62.01 & 56.80 \\
    \bottomrule
    \end{tabular}
    }
    \caption{Few-shot sentiment classification performance of open-source LLMs on RoIt-XMASA, showing performance by domain and language. Only baseline (0-shot) and best-performing few-shorts configurations are shown for each model.}
    \label{tab:llm_results}
\end{table*}

We present our experimental findings organized into three subsections: multi-adversarial training with multilingual encoders, few-shot performance of open-source LLMs, and the impact of unlabeled adaptive fine-tuning. All tables report performance disaggregated by domain and language, with the final column showing macro-averaged results.

\subsection{Multi-Adversarial Training Performance}

In this set of experiments, we apply our framework to three established multilingual models: M-BERT, XLM, and XLM-R. Table~\ref{tab:finetune_results} shows that our multi-target adversarial training framework consistently improves sentiment classification performance across all baseline models. The most significant gains are observed when both domain and language are used as adversarial objectives, confirming that learning representations invariant to both factors is beneficial.

XLM-R emerges as the top-performing model overall, achieving an average F1-score of 66.23\% with multi-adversarial training, a 4.64\% improvement over its baseline of 61.59\%. However, the model performance varies across domains and languages, and no single model dominates all settings. XLM-R excels at movies (68.61\% F1-score) and Italian reviews (69.40\% F1-score), while XLM-R leads both in book performance (64.12\% F1-score) and in music performance (65.78\% F1-score). Notably, M-BERT achieves the highest Romanian F1-score (61.35\%), outperforming both XLM and XLM-R in this more challenging language. These complementary strengths suggest that ensemble approaches or language-specific model selection could further improve cross-lingual sentiment analysis.

\subsection{LLM Baseline Performance}

The evaluation results, summarized in Table~\ref{tab:llm_results}, reveal the current capabilities of these models in a few-shot learning context for this task. Overall, performance improves with an increased number of in-context examples, although the gains are not always monotonic.

Llama-3.1-8B achieves the highest overall performance with a 58.43\% F1-score in the 5-shot setting, excelling particularly on movies (61.85\% F1-score). However, the model strengths vary by domain and language: DeepSeek-R1-8B demonstrates better performance on books (57.81\% F1-score) and Romanian reviews (57.04\% F1-score), while Qwen3-8B leads in music classification (59.34\% F1-score). The cross-lingual gap persists across all models, though DeepSeek-R1-8B shows the smallest Italian-Romanian disparity (61.23\% vs. 57.04\% F1-scores), suggesting better multilingual calibration.

These results highlight an important trade-off between the two paradigms. Although the best fine-tuned encoder (XLM-R with multi-adversarial training at 66.23\% F1-score) achieves notably higher performance than the best few-shot LLM (Llama-3.1-8B at 58.43\% F1-score), the LLM-based approach requires no task-specific labeled data or gradient-based optimization, relying solely on a handful of in-context examples. This observation makes a few-shot LLM evaluation particularly attractive in low-resource scenarios where the labeled dataset is scarce or annotation is costly. Conversely, when sufficient labeled data and computational resources are available, task-specific fine-tuning with adversarial training remains the more performant choice, underscoring the complementary nature of both approaches given practical constraints.

\subsection{Unlabeled Adaptive Fine-Tuning}

\begin{table*}
\centering
\small
\resizebox{\textwidth}{!}{
\begin{tabular}{|l|l|cc|cc|cc|cc|cc|cc|}
\toprule
\multirow{2}{*}{\textbf{Model}} & \multirow{2}{*}{\textbf{Config.}} & \multicolumn{2}{c|}{\textbf{Books}} & \multicolumn{2}{c|}{\textbf{Movies}} & \multicolumn{2}{c|}{\textbf{Music}} & \multicolumn{2}{c|}{\textbf{IT}} & \multicolumn{2}{c|}{\textbf{RO}} & \multicolumn{2}{c|}{\textbf{Avg.}} \\
\cmidrule{3-14}
& & \textbf{Acc.} & \textbf{F1} & \textbf{Acc.} & \textbf{F1} & \textbf{Acc.} & \textbf{F1} & \textbf{Acc.} & \textbf{F1} & \textbf{Acc.} & \textbf{F1} & \textbf{Acc.} & \textbf{F1} \\
\midrule
\multirow{2}{*}{Llama-3.1-8B} 
 & Base & 61.89 & 56.78 & 66.23 & 61.85 & 63.12 & 58.15 & 66.34 & 60.89 & 61.15 & 56.63 & 63.66 & 58.43 \\
 & Adapted & 63.45 & 58.92 & \textbf{68.12} & \textbf{63.78} & \textbf{65.28} & \textbf{60.45} & \textbf{67.89} & 62.34 & 63.34 & 59.43 & \textbf{65.22} & \textbf{60.15} \\
\midrule
\multirow{2}{*}{Qwen3-8B} 
 & Base & 59.67 & 54.23 & 61.45 & 56.12 & 63.78 & 59.34 & 63.45 & 57.92 & 59.82 & 55.11 & 60.11 & 54.97 \\
 & Adapted & 61.78 & 56.45 & 63.89 & 59.18 & 64.12 & 59.87 & 65.12 & 59.34 & \textbf{64.56} & \textbf{60.78} & 62.14 & 57.20 \\
\midrule
\multirow{2}{*}{Mistral-7B-v0.3} 
 & Base & 53.89 & 47.23 & 57.45 & 51.34 & 54.56 & 48.40 & 57.89 & 51.12 & 53.71 & 46.86 & 55.30 & 48.99 \\
 & Adapted & \textbf{56.23} & 49.78 & 58.12 & 51.89 & 55.21 & 49.45 & 58.34 & 51.67 & 55.40 & 48.78 & 56.85 & 50.12 \\
\midrule
\multirow{2}{*}{DeepSeek-R1-8B} 
 & Base & 62.45 & 57.81 & 63.34 & 59.12 & 61.23 & 55.67 & 65.12 & 61.23 & 61.89 & 57.04 & 62.01 & 56.80 \\
 & Adapted & 63.78 & \textbf{59.34} & 64.23 & 59.78 & 61.89 & 56.23 & 66.12 & \textbf{62.89} & 63.67 & 58.12 & 62.35 & 56.95 \\
\bottomrule
\end{tabular}
}

\caption{Impact of 1-epoch LoRA adaptation on LLMs, showing base vs. adapted performance by domain and language. Results are reported using the best-performing few-shot configuration for each model.}
\label{tab:unlabeled_adaptation}
\end{table*}

To evaluate the impact of domain-specific adaptation without labeled task data, we conducted a series of experiments using unlabeled adaptive fine-tuning. In this setup, the base LLMs were fine-tuned for a single epoch on the 202,141 unlabeled samples of the RoIt-XMASA corpus using LoRA. This process focuses exclusively on language modeling the target domains (books, movies, and music) in Romanian and Italian, without any exposure to sentiment labels.

The results in Table \ref{tab:unlabeled_adaptation} show consistent improvements in all models, although the adaptation benefits vary by model and domain. Llama-3.1-8B gains 1.72 percentage points, with the greatest improvements in movies (+1.93 in F1-score) and music (+2.30 in F1-score). Qwen3-8B shows the strongest increase for Romanian reviews (+5.67 in F1-score), suggesting a particularly effective cross-lingual adaptation from the unlabeled corpus. DeepSeek-R1-8B exhibits strong book adaptation (+1.53 in F1-score) and achieves the best Italian performance after adaptation (62.89\% in F1-score), while Mistral-7B-v0.3 shows modest but consistent gains across all domains. These varied adaptation patterns demonstrate that brief unsupervised exposure to target domains and languages measurably enhances LLM performance, with model-specific strengths emerging through domain alignment.

\section{Conclusion}
 
This paper introduced RoIt-XMASA, an extension of the XMASA framework to Italian and Romanian, using 36,000 labeled reviews across three domains. Our multi-target adversarial training framework with meta-learned coefficients achieved significant improvements, with XLM-R reaching 66.23\% F1-score when using both domain and language as adversarial targets. Few-shot LLM baselines, while lower in absolute performance, achieved competitive results without task-specific fine-tuning, revealing a practical trade-off between, on the one hand, labeling and training costs, and, on the other, classification accuracy. The dataset reveals distinct cross-linguistic patterns, with Romanian reviews averaging 2.4$\times$ longer than Italian reviews, underscoring the importance of accounting for linguistic variation in multilingual NLP systems. Future work should explore incorporating the unlabeled data through semi-supervised learning and extend the framework to additional low-resource languages.

\section*{Ethics Statement}
The RoIt-XMASA dataset was compiled from publicly available e-commerce reviews, in accordance with the terms of service of the source platforms. To protect user privacy, all personally identifiable information and author metadata were removed, leaving only review text, titles, and ratings. The dataset is released for academic research purposes only. Although rigorous cleaning was applied, the reviews were not filtered for toxic language to preserve natural linguistic distributions; consequently, the corpus may contain offensive content, typical of unmoderated user-generated text.

\section*{Acknowledgements}
This work was supported by the National University of Science and Technology POLITEHNICA Bucharest through the PubArt program.

\bibliography{aaai2026}

\section{Checklist}

\begin{enumerate}

\item For most authors...
\begin{enumerate}
  \item  Would answering this research question advance science without violating social contracts, such as violating privacy norms, perpetuating unfair profiling, exacerbating the socio-economic divide, or implying disrespect to societies or cultures?
    \answerYes{Yes, this work introduces a multilingual dataset to improve sentiment analysis for underrepresented languages such as Romanian, following established ethical scraping and cleaning protocols.}
  \item Do your main claims in the abstract and introduction accurately reflect the paper's contributions and scope?
    \answerYes{Yes, the contributions regarding the RoIt-XMASA dataset and the multi-target adversarial training framework are outlined in the abstract and introduction.}
   \item Do you clarify how the proposed methodological approach is appropriate for the claims made? 
    \answerYes{Yes, we describe the multi-adversarial objective and meta-learning strategy used to balance sentiment discrimination with domain/language invariance.}
   \item Do you clarify what are possible artifacts in the data used, given population-specific distributions?
    \answerYes{Yes, we discuss the rating distribution and textual characteristics, noting peaks at the extreme ratings and differences in review styles between Romanian and Italian.}
  \item Did you describe the limitations of your work?
    \answerYes{Yes, limitations related to resource scarcity for Romanian and the challenges of the double shift in the domain and language are discussed in the Introduction.}
  \item Did you discuss any potential negative societal impacts of your work?
    \answerYes{Yes, the paper mentions considerations regarding cultural nuances and potential model failures in capturing these subtleties.}
      \item Did you discuss any potential misuse of your work?
    \answerNo{No, because the primary application is sentiment analysis for e-commerce reviews, which has low potential for harmful misuse beyond standard concerns for all NLP models.}
    \item Did you describe steps taken to prevent or mitigate potential negative outcomes of the research, such as data and model documentation, data anonymization, responsible release, access control, and the reproducibility of findings?
    \answerYes{Yes, we detail rigorous cleaning, language verification, manual verification for quality, and provide a HuggingFace link for responsible data release.}
  \item Have you read the ethics review guidelines and ensured that your paper conforms to them?
    \answerYes{Yes, the data collection followed native linguistic patterns and language-specific acquisition strategies to ensure authentic and ethical data usage.}
\end{enumerate}

\item Additionally, if your study involves hypotheses testing...
\begin{enumerate}
  \item Did you clearly state the assumptions underlying all theoretical results?
    \answerNA{N/A}
  \item Have you provided justifications for all theoretical results?
    \answerNA{N/A}
  \item Did you discuss competing hypotheses or theories that might challenge or complement your theoretical results?
    \answerNA{N/A}
  \item Have you considered alternative mechanisms or explanations that might account for the same outcomes observed in your study?
    \answerNA{N/A}
  \item Did you address potential biases or limitations in your theoretical framework?
    \answerNA{N/A}
  \item Have you related your theoretical results to the existing literature in social science?
    \answerNA{N/A}
  \item Did you discuss the implications of your theoretical results for policy, practice, or further research in the social science domain?
    \answerNA{N/A}
\end{enumerate}

\item Additionally, if you are including theoretical proofs...
\begin{enumerate}
  \item Did you state the full set of assumptions of all theoretical results?
    \answerNA{N/A}
	\item Did you include complete proofs of all theoretical results?
    \answerNA{N/A}
\end{enumerate}

\item Additionally, if you ran machine learning experiments...
\begin{enumerate}
  \item Did you include the code, data, and instructions needed to reproduce the main experimental results (either in the supplemental material or as a URL)?
    \answerYes{Yes, the dataset is released on HuggingFace (see Footnote 1).}
  \item Did you specify all the training details (e.g., data splits, hyperparameters, how they were chosen)?
    \answerYes{Yes, we specify splits in Table 1 and provide full hyperparameter details in Appendices.}
     \item Did you report error bars (e.g., with respect to the random seed after running experiments multiple times)?
    \answerNo{No, because single-run results were reported for benchmarking; however, a fixed random seed (42) was used for reproducibility.}
	\item Did you include the total amount of compute and the type of resources used (e.g., type of GPUs, internal cluster, or cloud provider)?
    \answerNo{No, because the focus was on F1-score performance and methodological novelty, though mixed-precision training was noted to reduce memory.}
     \item Do you justify how the proposed evaluation is sufficient and appropriate to the claims made? 
    \answerYes{Yes, we evaluate across three domains and two languages using both encoder-based and LLM baselines to validate our adversarial framework.}
     \item Do you discuss what is ``the cost`` of misclassification and fault (in)tolerance?
    \answerNo{No, as the task is four-class sentiment analysis, where errors represent deviations in user perception rather than critical system failure.}
  
\end{enumerate}

\item Additionally, if you are using existing assets (e.g., code, data, models) or curating/releasing new assets, \textbf{without compromising anonymity}...
\begin{enumerate}
  \item If your work uses existing assets, did you cite the creators?
    \answerYes{Yes, we cite the original XMASA dataset creators and the developers of models like XLM-R and Llama-3.1.}
  \item Did you mention the license of the assets?
    \answerNo{No, because the models used (Llama, XLM-R) are under well-known open-source licenses cited in the references.}
  \item Did you include any new assets in the supplemental material or as a URL?
    \answerYes{Yes, the RoIt-XMASA dataset is available via the provided HuggingFace link.}
  \item Did you discuss whether and how consent was obtained from people whose data you're using/curating?
    \answerYes{Yes, data was collected from public e-commerce platforms using language-specific acquisition strategies.}
  \item Did you discuss whether the data you are using/curating contains personally identifiable information or offensive content?
    \answerYes{Yes, manual verification by native speakers assessed text quality and alignment, ensuring samples represent unique review instances.}
\item If you are curating or releasing new datasets, did you discuss how you intend to make your datasets FAIR (see \citet{fair})?
\answerNo{No, though the dataset is publicly hosted on HuggingFace to ensure accessibility.}
\item If you are curating or releasing new datasets, did you create a Datasheet for the Dataset (see \citet{gebru2021datasheets})? 
\answerNo{No, but detailed statistical characteristics are provided in Section 3 and Appendices.}
\end{enumerate}

\item Additionally, if you used crowdsourcing or conducted research with human subjects, \textbf{without compromising anonymity}...
\begin{enumerate}
  \item Did you include the full text of instructions given to participants and screenshots?
    \answerNA{N/A}
  \item Did you describe any potential participant risks, with mentions of Institutional Review Board (IRB) approvals?
    \answerNA{N/A}
  \item Did you include the estimated hourly wage paid to participants and the total amount spent on participant compensation?
    \answerNA{N/A}
   \item Did you discuss how data is stored, shared, and deidentified?
   \answerYes{Yes, we applied cleaning and normalization to ensure data quality and integrity.}
\end{enumerate}

\end{enumerate}

\appendix

\section{Cross-Language Analysis}
\label{app:cross_language}

\begin{figure*}
\centering
\includegraphics[width=\textwidth]{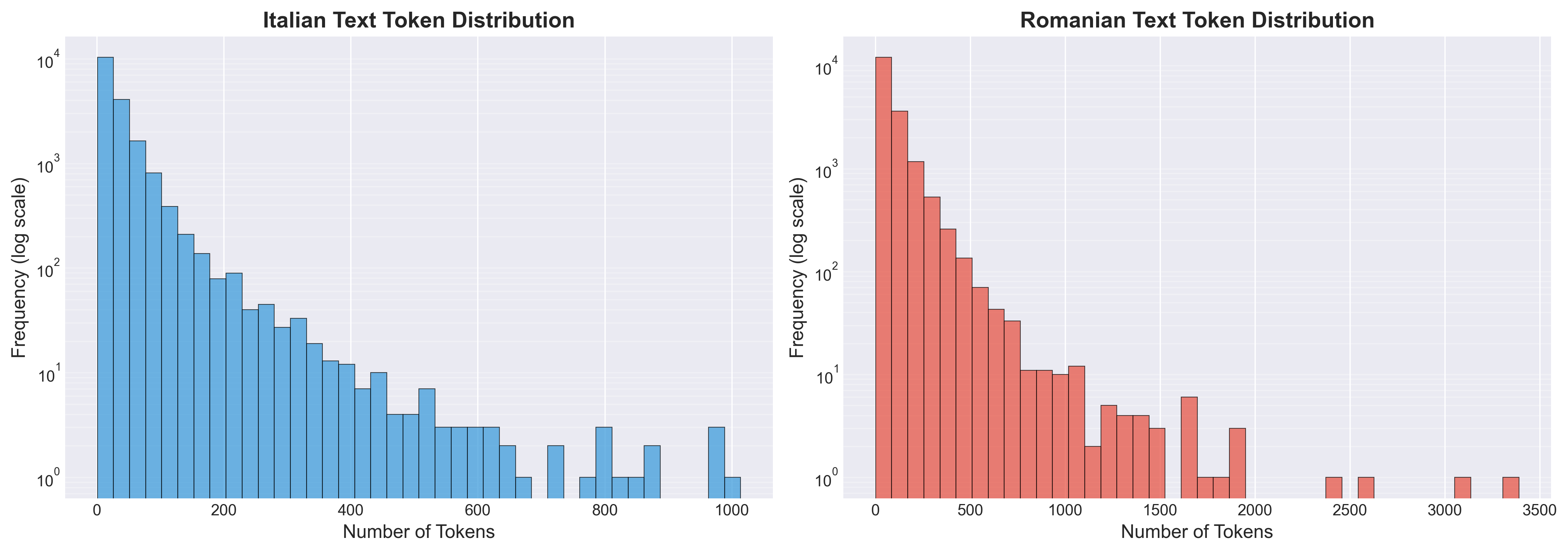}
\caption{Token distribution of the text in RoIt-XMASA, grouped by language.}
\label{fig:text_hist_lang}
\end{figure*}

The RoIt-XMASA dataset reveals differences in the reviewing patterns between Italian and Romanian users. Romanian reviews are, on average, 2.4 times longer than Italian reviews (89.02 vs. 37.34 tokens), suggesting more elaborate review styles in Romanian online communities. This pattern is clearly visible in Figure~\ref{fig:text_hist_lang}, where the Romanian distribution shows a longer tail with reviews that frequently exceed 500 tokens, while Italian reviews concentrate below 200 tokens.

\begin{figure*}
\centering
\includegraphics[width=\textwidth]{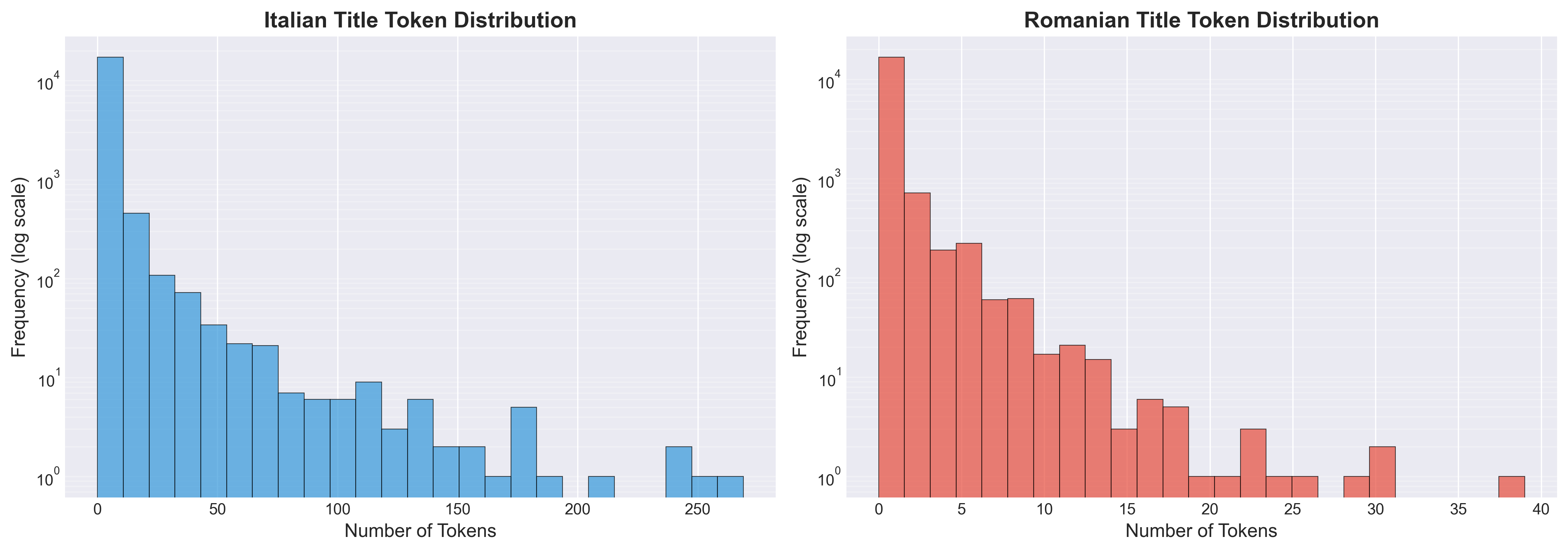}
\caption{Token distribution of the title in RoIt-XMASA, grouped by language.}
\label{fig:title_hist_lang}
\end{figure*}

The usage patterns of the titles show an inverse relationship, as illustrated in Figure~\ref{fig:title_hist_lang}. Italian reviews show a strong preference for including titles, whereas Romanian reviews rarely include titles. The Italian distribution shows clear peaks at 2-5 tokens, suggesting standardized title formats, while the Romanian distribution is heavily concentrated at zero, indicating different platform conventions or user behaviors.

These language-specific patterns in RoIt-XMASA have important implications for model architecture choices, particularly regarding maximum sequence length and attention mechanisms. The substantial difference in review lengths suggests that models may benefit from language-specific pre-processing or architectural adaptations.

\section{Cross-Domain Analysis}
\label{app:cross_domain}

\begin{figure*}
\centering
\includegraphics[width=\textwidth]{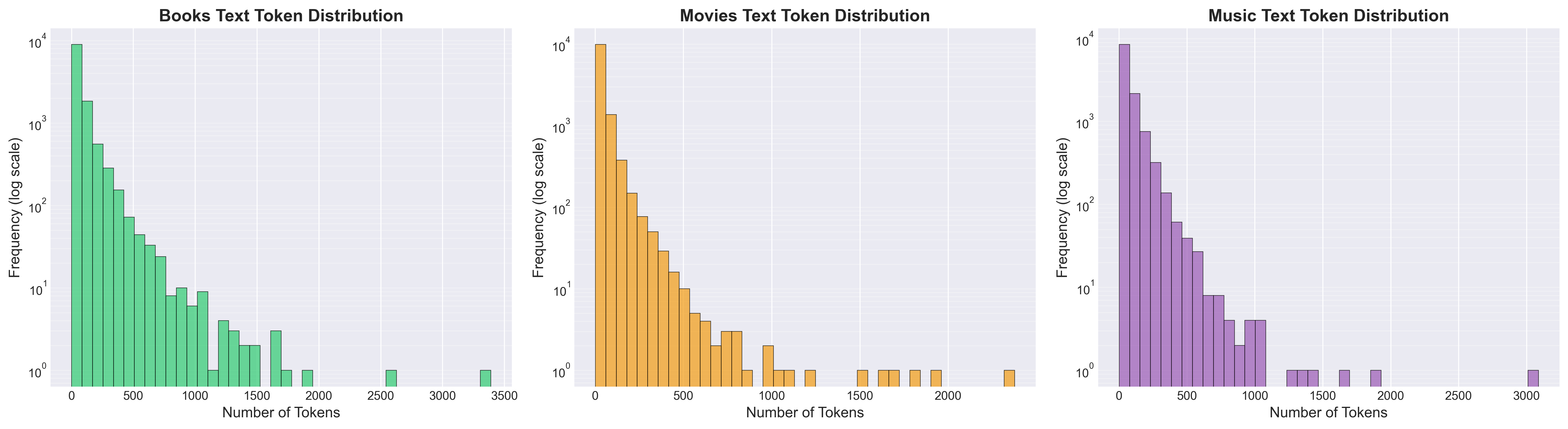}
\caption{Token distribution of the text in RoIt-XMASA, grouped by domain.}
\label{fig:text_hist_domain}
\end{figure*}

Domain-specific review patterns emerge clearly from the RoIt-XMASA token distributions shown in Figure~\ref{fig:text_hist_domain}. Book reviews have the highest mean length (76.27 tokens), with a broad distribution that extends beyond 1,000 tokens, reflecting detailed literary analysis and plot discussion. Music reviews follow closely (71.53 tokens) with similar variance, while movie reviews are notably more concise (41.73 tokens) with a distribution concentrated below 500 tokens.

\begin{figure*}
\centering
\includegraphics[width=\textwidth]{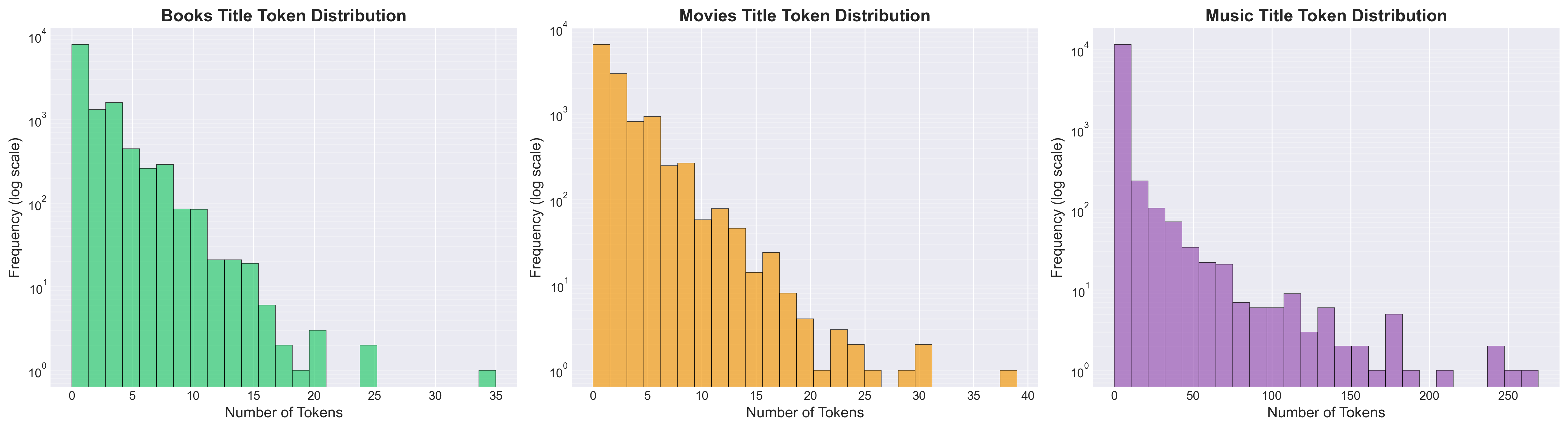}
\caption{Token distribution of the title in RoIt-XMASA, grouped by domain.}
\label{fig:title_hist_domain}
\end{figure*}

The usage of titles varies by domain, as shown in Figure~\ref{fig:title_hist_domain}. Music reviews exhibit the highest variance in title length (std: 11.12), with a bimodal distribution showing peaks at both 0 and 50+ tokens, reflecting users who either omit titles entirely or include full album/track listings. Book and movie reviews show more consistent usage patterns of titles with distributions concentrated below 20 tokens.

Interesting interaction effects emerge when examining language-domain combinations in RoIt-XMASA. The Italian-Romanian length ratio ranges from 1.6:1 in movie reviews to 3.7:1 in music reviews, indicating that cultural factors influence discussions of different media types in each language. These substantial cross-domain differences validate the continued relevance of the domain adaptation challenges identified by \citet{blitzer2007biographies}, now extended to multilingual settings through the RoIt-XMASA dataset.

\section{Text Normalization Pipeline}
\label{app:normalization}
The following normalization rules were applied to all reviews in the RoIt-XMASA dataset to ensure consistent text representation for transformer-based models 
\cite{vaswani2017attention} while preserving essential, semantic, and syntactic information for sentiment analysis:

\begin{enumerate}
\item \textbf{Punctuation normalization}: Multiple consecutive punctuation marks were reduced to a maximum of three instances (e.g., "!!!!!" → "!!!"), preserving emphasis while preventing excessive repetition. This includes ellipses ("......" → "...") and mixed punctuation patterns.
\item \textbf{Whitespace standardization}: All sequences of multiple whitespace characters (spaces, tabs, and non-breaking spaces) were replaced with single spaces, while leading and trailing whitespaces were removed from each review.
\item \textbf{URL and email handling}: URLs and email addresses were replaced with special tokens \texttt{[URL]} and \texttt{[EMAIL]}, respectively, preserving the information that external references existed, while removing potentially noisy string patterns.
\end{enumerate}

These normalization steps were designed to balance data cleaning with information preservation, ensuring optimal performance for transformer-based models while maintaining the linguistic characteristics essential for sentiment analysis in Italian and Romanian.

\section{Hyperparameters for Multilingual Encoder Models}
\label{app:bert_hyperparams}

Table~\ref{tab:hyperparams} outlines the hyperparameters used to fine-tune M-BERT, XLM, and XLM-R for sentiment classification tasks. We employed early stopping with a patience of 3 epochs based on the validation set F1-score to prevent overfitting. The meta-learning rate $\eta$ for the adversarial coefficient updates was set to 0.01, with the coefficients initialized to $\lambda_1 = \lambda_2 = 0.5$ and reduced in the range $[0, 2]$ to ensure training stability. The adversarial coefficients were updated every 100 training steps using the meta-learning procedure described in Section 4.1.

All experiments used mixed-precision training (FP16) to reduce memory consumption and accelerate training. We set the random seed to 42 for reproducibility across all runs. The classification head consisted of a single linear layer with dropout applied before the final projection. For adversarial discriminators (i.e., domain and language), we used identical single-layer architectures with the same dropout rate.

\begin{table}
    \centering
    \begin{tabular}{|l|c|}
    \toprule
    \textbf{Hyperparameter} & \textbf{Value} \\
    \midrule
    Optimizer & AdamW \\
    Learning Rate & 2e-5 \\
    Batch Size & 32 \\
    Training Epochs & 3--5 \\
    Max Sequence Length & 128 \\
    Dropout Probability & 0.1 \\
    Weight Decay & 0.01 \\
    Warmup Ratio & 0.1  \\
    Gradient Clipping & 1.0 \\
    Meta Learning Rate ($\eta$) & 0.01 \\
    Early Stopping Patience & 3 \\
    Random Seed & 42 \\
    \bottomrule
    \end{tabular}
    \caption{Hyperparameters used for fine-tuning multilingual encoder models.}
    \label{tab:hyperparams}
\end{table}

\section{Hyperparameters for LLMs}
\label{app:llm_hyperparams}

This section details the configuration for the LLM baselines. During the evaluation phase, we used greedy decoding (setting the temperature to 0.0) to ensure deterministic, reproducible sentiment-rating predictions. The inference process used a maximum limit of 5 for new tokens, which was sufficient to generate a single-number rating.

The unsupervised domain adaptation phase focused on the 202,141 unlabeled reviews provided in the RoIt-XMASA dataset. This stage used the AdamW optimizer \cite{loshchilovdecoupled} with a cosine learning rate scheduler. The value-based hyperparameters for the LoRA adaptation are provided in Table~\ref{tab:llm_hyperparams}.

\begin{table}
\centering
\small
\begin{tabular}{|l|r|}
\toprule
\textbf{Hyperparameter} & \textbf{Value} \\
\midrule
Epochs & 1 \\
LoRA Rank ($r$) & 16 \\
LoRA Alpha ($\alpha$) & 32 \\
LoRA Dropout & 0.05 \\
Learning Rate & 2e5 \\
Warmup Ratio & 0.05 \\
Weight Decay & 0.01 \\
Global Batch Size & 16 \\
\bottomrule
\end{tabular}
\caption{Hyperparameter settings for the LLM unsupervised adaptation and inference phases.}
\label{tab:llm_hyperparams}
\end{table}

\section{LLM Evaluation Prompts}
\label{app:llm_prompt}

In this section, we present the templates used for the zero-shot and few-shot evaluations of the LLMs. The templates provided below are the translated versions used for querying the models (i.e., from Romanian and Italian); for the few-shot configurations, the placeholders for \texttt{Title}, \texttt{Review}, and \texttt{Rating} were populated with the respective language-specific samples from the RoIt-XMASA dataset.

\begin{flushleft}
\begin{minipage}{0em}
\fbox{\begin{minipage}{21.15em}
Zero-Shot
\end{minipage}}
\fbox{\begin{minipage}{21.15em}
You are a review rating predictor. Given a review text, predict its rating on a scale of 1 to 5 (except 3).\\

1 = Very negative \\
2 = Negative \\
4 = Positive \\
5 = Very positive \\

Only respond with a single number (1, 2, 4, or 5).

Title: \{\} \\
Review: \{\} \\
Rating:
\end{minipage}}
\end{minipage}
\end{flushleft}

\begin{flushleft}
\begin{minipage}{0em}
\fbox{\begin{minipage}{21.15em}
Multi-Shot
\end{minipage}}
\fbox{\begin{minipage}{21.15em}
You are a review rating predictor. Given a review text, predict its rating on a scale of 1 to 5 (except 3).\\

1 = Very negative \\
2 = Negative \\
4 = Positive \\
5 = Very positive \\

Only respond with a single number (1, 2, 4, or 5). \\

Here are some examples:
\\
Title1: \{\} \\
Review1: \{\} \\
Rating1: \{\} \\
...
\\
\\
Now predict the rating for this review: \\
Title: \{\} \\
Review: \{\} \\
Rating:
\end{minipage}}
\end{minipage}
\end{flushleft}

\end{document}